\documentclass{article}

\usepackage{microtype}
\usepackage{graphicx}
\usepackage{subcaption}
\usepackage{booktabs} 

\usepackage{hyperref}

 \usepackage[preprint]{icml2026}

\usepackage{amsmath}
\usepackage{amssymb}
\usepackage{mathtools}
\usepackage{amsthm}
\usepackage{array}
\usepackage{multirow}
\usepackage{booktabs}
\usepackage{makecell}
\usepackage{adjustbox}

\usepackage[capitalize,noabbrev]{cleveref}

\theoremstyle{plain}

\theoremstyle{definition}

\theoremstyle{remark}

\usepackage[textsize=tiny]{todonotes}

\icmltitlerunning{Rethinking Efficient Mixture-of-Experts for Remote Sensing Modality-Missing Classification}

\begin{document}

\twocolumn[
\icmltitle{Rethinking Efficient Mixture-of-Experts for Remote Sensing Modality-Missing Classification}

\begin{icmlauthorlist}
	\icmlauthor{Qinghao Gao}{xdu}
	\icmlauthor{Jiahui Qu}{xdu}
	\icmlauthor{Wenqian Dong}{xdu}
\end{icmlauthorlist}

\icmlaffiliation{xdu}{
	State Key Laboratory of Integrated Service Network,\\
	Xidian University, Xi'an 710071, China
}

\icmlcorrespondingauthor{Qinghao Gao}{qhgao@stu.xidian.edu.cn}

\icmlkeywords{Remote Sensing, Modality Missing, Mixture-of-Experts, ICML}

\vskip 0.3in
]



 \printAffiliationsAndNotice{\icmlEqualContribution}

\begin{abstract}
	
Multimodal remote sensing classification often suffers from missing modalities caused by sensor failures and environmental interference, leading to severe performance degradation. In this work, we rethink missing-modality learning from a conditional computation perspective and investigate whether Mixture-of-Experts (MoE) models can inherently adapt to diverse modality-missing scenarios. We first conduct a systematic study of representative MoE paradigms under various missing-modality settings, revealing both their potential and limitations. Building on these insights, we propose a Missing-aware Mixture-of-LoRAs (MaMOL), a parameter-efficient MoE framework that unifies multiple modality-missing cases within a single model. MaMOL introduces a dual-routing mechanism to decouple modality-invariant shared experts and modality-aware dynamic experts, enabling automatic expert activation conditioned on available modalities. Extensive experiments on multiple remote sensing benchmarks demonstrate that MaMOL significantly improves robustness and generalization under diverse missing-modality scenarios with minimal computational overhead. Transfer experiments on natural image datasets further validate its scalability and cross-domain applicability.

\end{abstract}

\section{Introduction}
\label{sec:intro}

Multimodal remote sensing has become a cornerstone for a wide range of Earth observation applications~\cite{liu2025cross, zhang2025haf, wei2024gradient,wang2025uncertainty,zhang2025cdprompt}, including land-cover classification, change detection, and object recognition. By leveraging the complementary properties of different sensing modalities, such as optical imagery, SAR, hyperspectral data, and LiDAR, multimodal models can learn richer representations and exhibit stronger robustness. However, in real-world remote sensing scenarios, the availability of all modalities cannot be guaranteed. Sensor failures, atmospheric conditions, limited acquisition budgets, and inconsistent coverage frequently lead to missing modalities~\cite{huangshijie, yang2025DPMamba}. Such incompleteness not only degrades performance but also fundamentally challenges conventional multimodal learning frameworks, which typically assume full-modality availability during training and inference. Moreover, missing modalities arise in diverse combinations, making remote sensing a particularly suitable testbed for evaluating the robustness and adaptability of multimodal models.

To address missing modalities in remote sensing, most existing methods rely on a two-stage paradigm~\cite{chen2024novel,wei2023diversity,wei2023msh}. A multimodal model is first trained with complete data, and then adapted to missing-modality scenarios through additional modules, retraining, or parameter updates. While these approaches have achieved encouraging results, they suffer from two major limitations. First, repeated retraining or adaptation incurs substantial computational and storage overhead, especially when facing multiple missing-modality patterns. Second, the reliance on complete data during initial training fails to account for the inherent incompleteness of real-world observations, limiting generalization to unseen modality-missing conditions. These limitations indicate the need to rethink missing-modality learning from a more fundamental perspective.

Recent advances in large foundation models for natural images and vision–language tasks provide new insights. Such models demonstrate that a single backbone can efficiently adapt to diverse input conditions through low-parameter prompt-based tuning, without requiring separate retraining for each scenario~\cite{lee2023cvpr,hu2024deep}. Although effective, these approaches typically require configuring specific prompts for each modality subset, causing the number of configurations to grow exponentially with the number of modalities. Moreover, the introduction of additional prompt branches increases architectural complexity and may hinder scalability in practical multi-sensor systems.

Motivated by these observations, we revisit missing-modality learning from a conditional computation perspective. Under this view, each available modality combination corresponds to a distinct computation path within the network, and the model should dynamically select appropriate computation strategies based on the observed inputs. Importantly, the core challenge is not only to compensate for missing information, but to adapt the internal computation process to the reliability and semantics of the available modalities. This perspective exposes structural properties of the missing-modality problem that remain largely underexplored in existing multimodal remote sensing frameworks.

\begin{figure}[t]
	\centering
	\includegraphics[width=1\linewidth]{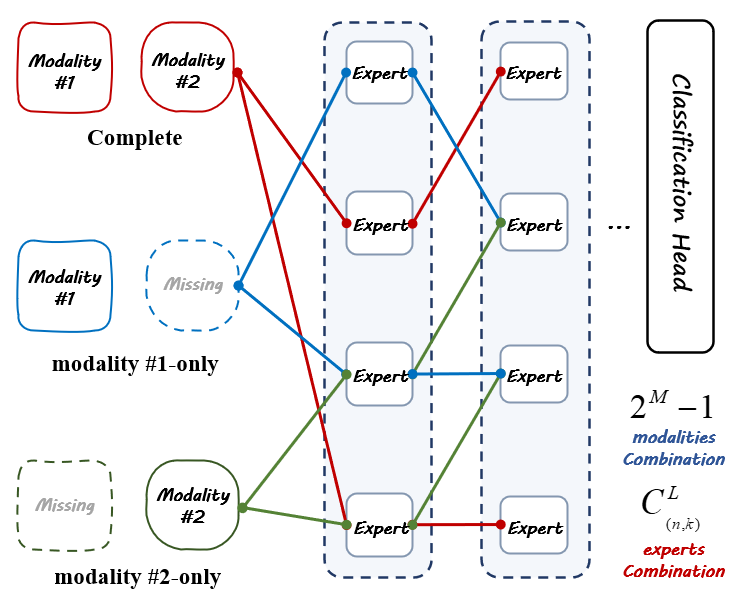}
	\caption{
		The MoE-based approach processes all possible input combinations from $M$ modalities through a single network containing $L$ modality expert layers. Each layer activates $k$ out of $n$ experts, forming $C_{(n,k)}^L$ data flow pathways through expert combinations to handle $2^M - 1$ missing-modality scenarios. Colored arrows indicate different modality inputs: red (complete), blue (modality \#1-only), green (modality \#2-only). This architecture achieves exponential combination capacity while maintaining a unified structure.}
	\label{fig:moe_framework}
\end{figure}

This structural insight naturally points to Mixture-of-Experts (MoE) models\cite{shazeer2017outrageously, riquelme2021scaling, fedus2022switch}
. MoE architectures consist of multiple expert subnetworks and a routing mechanism that activates experts conditioned on the input. Such designs are inherently compatible with missing-modality scenarios, where different modality subsets can be interpreted as different input conditions requiring specialized computation paths.As illustrated in Fig.~\ref{fig:moe_framework}, an MoE-based approach can process all $2^M - 1$ possible modality combinations through a single network with $L$ expert layers, achieving exponential combination capacity while maintaining a unified structure. However, existing MoE formulations are mainly developed for specific tasks or domains, without explicitly considering modality-aware routing. As a result, how to effectively exploit MoE models for missing-modality remote sensing classification remains unclear. In this work, we therefore systematically investigate a range of representative MoE variants, analyzing their robustness, adaptability, and parameter efficiency under diverse missing-modality patterns.

Building upon these explorations, we further propose a Modality-aware Mixture of LoRA Experts (MaMOL), a parameter-efficient MoE framework that unifies multiple missing-modality scenarios within a single model. MaMOL introduces a dual-routing design: modality-invariant shared experts preserve global cross-modal knowledge, while task-oriented dynamic experts are selectively activated according to the observed modality subset. By decoupling shared and dynamic experts, MaMOL avoids separate retraining and excessive parameter updates, enabling efficient adaptation to arbitrary missing-modality patterns.
Our main contributions are threefold:
\begin{itemize}
	\item We reformulate missing-modality learning in remote sensing as a conditional expert selection problem and present the first systematic study of MoE paradigms under missing-modality settings.
	\item We propose MaMOL, a dual-routing, parameter-efficient MoE framework based on lightweight LoRA experts, capable of handling diverse missing-modality scenarios within a single unified model.
	\item Extensive experiments on multiple remote sensing benchmarks demonstrate the robustness and generalization of MaMOL, and further results on natural image datasets validate its scalability beyond the remote sensing domain.
\end{itemize}

\section{Related Work}
\label{sec:formatting}

\subsection{Missing-Modality for Multimodal Learning}

Multimodal learning has achieved remarkable success across various downstream tasks~\cite{guo2024skysense,costanzino2024multimodal,Cheng_2025_CVPR,Swetha_2023_ICCV,Zhao_2023_ICCV,Yamaguchi_2025_CVPR}. However, incomplete or missing modalities often lead to severe performance degradation. Early studies focused on evaluating the robustness of multimodal Transformers under missing-modality conditions~\cite{ma2022multimodal,Huang_2025_ICCV,Pipoli_2025_ICCV}. Meanwhile, several works~\cite{zeng2022tag,Ke_2025_CVPR,Zhuang_2025_ICCV,chen2024probabilistic} mitigated this issue by reconstructing or predicting missing modality features, either in data or latent space. Although effective, such methods are computationally expensive and perform poorly with heterogeneous modalities~\cite{Dai_2025_ICCV,eccv2}. Recently, prompt-based parameter-efficient fine-tuning (PEFT) has emerged as a lightweight and effective solution~\cite{lee2023cvpr,hu2024deep,eccv_hanguo,Zhang_2025_ICCV,zhang2025cdprompt,yang2025DPMamba}. MAP~\cite{lee2023cvpr} assigning distinct prompt vectors for each missing-modality configuration while tuning less than 1\% of parameters. Hu~\cite{hu2024deep} further proposed a feature-driven dynamic prompt generation strategy that decomposes prompts into modality-shared and modality-specific components to exploit cross-modal complementarity.

\begin{figure*}[h]
	\centering
	\includegraphics[width=0.95\linewidth]{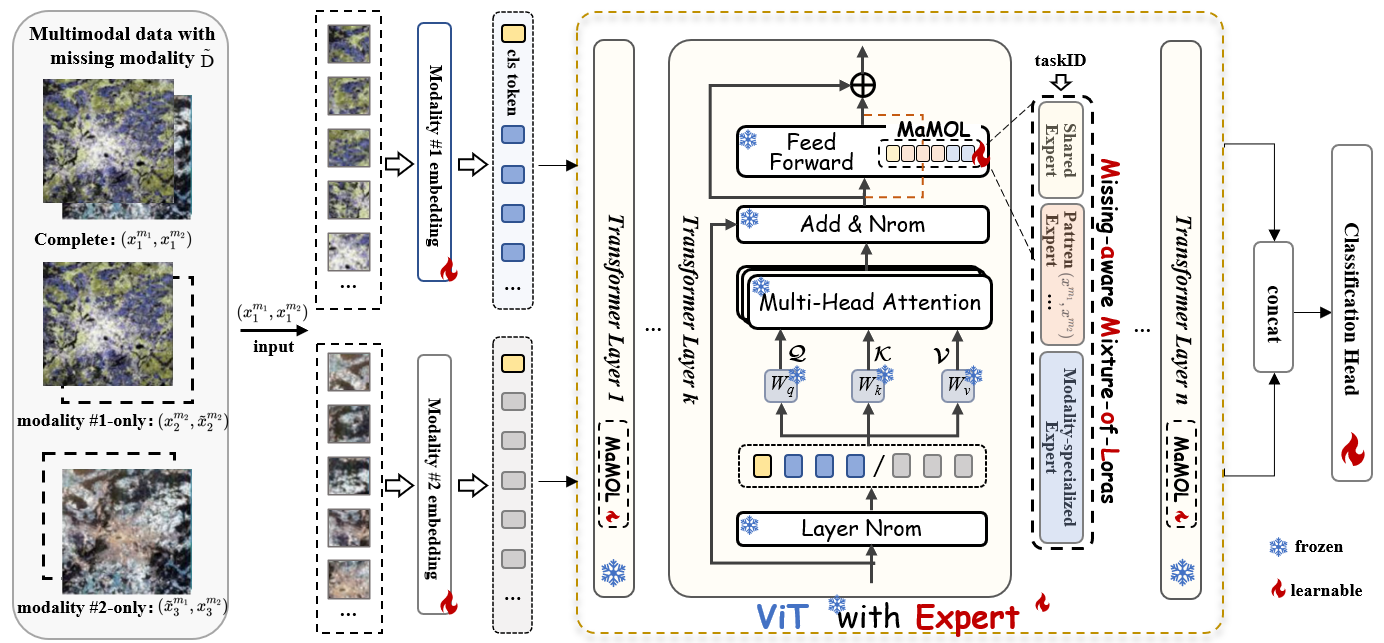}
	\caption{
		Overall architecture of the proposed framework.
		Multimodal data with potential missing modalities are embedded into modality-specific representations.
		These embeddings are then fed into a Transformer backbone enhanced with a Missing-aware Mixture-of-Loras (MaMOL) module.
		The design includes shared experts, pattern experts, and modality-specialized experts, enabling both general knowledge transfer and modality-aware adaptation.
		Finally, the aggregated features are concatenated and passed into the classification head.
	}
	\label{fig:framework}
\end{figure*}
\subsection{Mixture-of-Experts}

The Mixture-of-Experts (MoE) architecture has achieved remarkable success across various domains, including natural language processing (NLP)~\cite{liu2025moiraimoe,Lu_2024_CVPR,Cong_2023_ICCV}, computer vision~\cite{Yu_2024_CVPR,11175038,Rahman_2025_WACV} and multimodal fusion~\cite{Mamedov_2025_WACV,li2025uni,Lee_2025_WACV}. Traditional dense MoE models incur substantial computational costs since all experts are activated simultaneously~\cite{Cong_2023_ICCV,Lee_2025_WACV}. In contrast, sparse MoE approaches~\cite{Yang_2024_CVPR,Wu_2024_CVPR} activate only a subset of experts for each input, significantly improving scalability and efficiency. In the field of computer vision, MoE has been introduced to enhance modularity and reduce computational redundancy~\cite{Yang_2024_CVPR,Yu_2024_CVPR}. In multi-task learning (MTL), MoE enables task-specific routing, allowing each task to select its specialized experts~\cite{Zhu_2024_CVPR,Ye_2023_ICCV}. Furthermore, combining MoE with instruction tuning has been shown to further enhance task adaptability~\cite{11230232,Zhu_2024_CVPR}. Models such as FuseMoE~\cite{han2024fusemoe} and Flex-MoE~\cite{yun2024flexmoe} utilize modality-specific dynamic routing to flexibly assign experts, effectively alleviating issues of modality imbalance and missing modalities. In our study, we further introduce MoE as a multi-task learner for the missing-modality problem, aiming to achieve more generalized and adaptive multimodal learning under incomplete input conditions.

\section{Preliminaries}
\label{sec:prelim}

\subsection{Problem Definition}
\label{sec:problem}

We study multimodal remote sensing classification under generalized missing-modality conditions. 
Let a multimodal observation be denoted as
\begin{equation}
	\mathbf{x} = \{x^{(1)}, x^{(2)}, \dots, x^{(M)}\}, \quad y \in \mathcal{Y},
\end{equation}
where $M$ is the number of modalities and $y$ is the ground-truth class label.

In real-world scenarios, only a subset of modalities $\mathcal{S} \subseteq \{1,\dots,M\}$ is available. We define a binary modality indicator:
\begin{equation}
	\mathbf{m} \in \{0,1\}^{M}, \quad m_i = \mathbb{I}(i \in \mathcal{S}).
\end{equation}

To preserve unified network inputs, missing modalities are replaced with dummy tokens $\tilde{x}^{(i)}$, yielding:
\begin{equation}
	\hat{x}^{(i)} = m_i x^{(i)} + (1 - m_i)\tilde{x}^{(i)}, \quad 
	\tilde{\mathbf{x}} = \{\hat{x}^{(1)}, \dots, \hat{x}^{(M)}\}.
\end{equation}

Each modality is encoded independently:
\begin{equation}
	\mathbf{h}^{(i)} = \mathcal{E}^{(i)}(\hat{x}^{(i)}), \quad 
	\mathbf{h}_0 = \Phi(\mathbf{h}^{(1)}, \dots, \mathbf{h}^{(M)}),
\end{equation}
where $\mathcal{E}^{(i)}$ denotes a modality-specific encoder and $\Phi(\cdot)$ is a fusion operator.

The learning objective is:
\begin{equation}
	\min_{\theta} \mathbb{E}_{(\tilde{\mathbf{x}}, \mathbf{m}, y)} 
	\left[ \mathcal{L}(f_\theta(\mathbf{h}_0, \mathbf{m}), y) \right],
\end{equation}
where a single model $f_\theta$ is expected to operate robustly under arbitrary missing-modality patterns.

\subsection{Mixture-of-Experts for Missing-Modality Learning}
\label{sec:moe_prelim}

A general Mixture-of-Experts (MoE) layer maps an input representation $\mathbf{z}\in\mathbb{R}^d$ to:
\begin{equation}
	\mathrm{MoE}(\mathbf{z}) = \sum_{k=1}^{K} g_k(\mathbf{z}) f_k(\mathbf{z}), \quad 
	\mathbf{g} = \mathrm{softmax}(R(\mathbf{z})),
\end{equation}
where $f_k$ denotes the $k$-th expert and $R(\cdot)$ is a routing function. In missing-modality learning, models must operate under exponentially many modality-availability patterns, where the key challenge is not only representation learning, but dynamically adapting the computation pathway to observed modalities. MoE provides a natural framework by enabling conditional expert activation, making it particularly suitable for handling diverse missing-modality conditions. Existing multimodal MoE methods can be broadly categorized into three paradigms: 
(i) replace-based MoE; 
(ii) adapt-based MoE; 
and (iii) task-driven MoE.
We review these paradigms and analyze their limitations from a unified perspective.

\subsubsection{Replace-based MoE}
\label{sec:replace}

Replace-based MoE architectures fully substitute backbone layers with expert blocks:
\begin{equation}
	\mathbf{h} = \sum_{k=1}^{K} g_k(\mathbf{z}) \mathcal{F}_k(\mathbf{z}),
\end{equation}
where each $\mathcal{F}_k$ is a complete feed-forward network.

Although expressive, such designs incur $\mathcal{O}(K|\mathcal{F}|)$ parameter growth, where each expert instantiates a full functional module, and require allocating experts to different missing-modality patterns, replace-based MoE quickly becomes impractical for multimodal remote sensing scenarios.

\subsubsection{Adapt-based MoE}
\label{sec:adapt}

Adapt-based MoE preserves a shared backbone and attaches expert residual branches:
\begin{equation}
	\mathbf{h} = \mathcal{F}_0(\mathbf{z}) + \sum_{k=1}^{K} g_k(\mathbf{z}) f_k(\mathbf{z}),
\end{equation}
where $\mathcal{F}_0$ is frozen or lightly tuned. This formulation significantly improves parameter efficiency. However, without explicitly modeling modality availability, the experts entangle modality priors with missing-pattern adaptations, leading to unstable specialization and limited generalization.

\subsubsection{Task-driven MoE}
\label{sec:task}

Task-driven MoE incorporates explicit task descriptors:
\begin{equation}
	\mathbf{g} = \mathrm{softmax}(R([\mathbf{z}; \mathbf{t}])), \quad \mathbf{t}=\psi(\mathbf{m}),
\end{equation}
where $\mathbf{t}$ encodes the missing-modality configuration. While this enables coarse task awareness, all missing scenarios are still mapped to a single expert space without structural decomposition, causing interference between modality-invariant, modality-specific, and missing-pattern transformations.

\subsubsection{Discussion: Why Existing MoE Is Insufficient}
\label{sec:discussion}

Existing MoE paradigms lack a structural alignment with the intrinsic factorization. They either replace the backbone entirely, entangle multiple semantic roles within a single expert pool, or rely on shallow task conditioning. Consequently, they fail to (1) explicitly preserve modality-invariant knowledge, (2) maintain modality-specific priors under missing inputs, and (3) isolate missing-pattern-induced adaptations. This motivates a missing-aware MoE design.

\section{Missing-aware Mixture-of-LoRAs}
\label{sec:mamol}

From the perspective of missing-modality learning, the ideal transformation can be factorized as:
\begin{equation}
	f(\mathbf{z}, \mathbf{m}) = f_{\text{shr}}(\mathbf{z}) + f_{\text{mod}}(\mathbf{z}, \mathbf{m}) + f_{\text{pat}}(\mathbf{z}, \mathbf{m}),
	\label{eq:factor}
\end{equation}
where $f_{\text{shr}}$ models modality-invariant knowledge, $f_{\text{mod}}$ preserves modality-dependent priors, and $f_{\text{pat}}$ captures missing-pattern-specific adaptations.

\begin{figure*}[t]
	\centering
	\includegraphics[width=\linewidth]{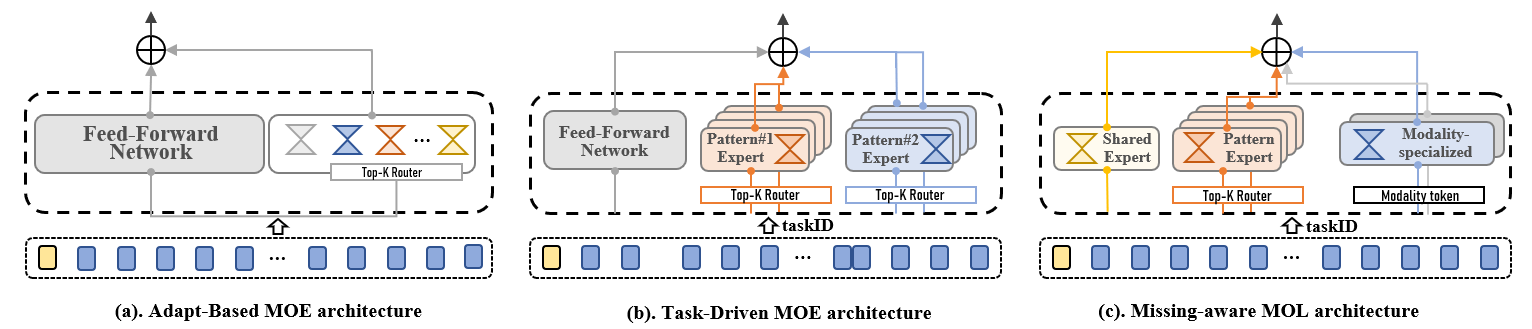}
	\caption{
		Comparison of different Mixture-of-Experts (MoE) architectures.
		(a) Adapt-Based MoE architecture.
		(b) Task-Driven MoE architecture.
		(c) Missing-aware Mixture-of-Loras (MaMOL) architecture.
	}
	\label{fig:3}
\end{figure*}

\subsection{Overall Framework}

As illustrated in Fig.~\ref{fig:framework}, we adopt a pretrained Vision Transformer (ViT) backbone with frozen parameters to minimize training cost. Let the $\ell$-th transformer block be:
\begin{equation}
	\mathbf{h}_\ell = \mathcal{T}_\ell(\mathbf{h}_{\ell-1}).
\end{equation}

To enable efficient adaptation under arbitrary modality-missing configurations, we propose Missing-aware Mixture-of-LoRAs (MaMOL), which injects low-rank expert residuals into the feed-forward layers:
\begin{equation}
	\tilde{\mathbf{h}}_\ell = \mathbf{h}_\ell + \Delta\mathbf{h}_\ell, \quad
	\Delta\mathbf{h}_\ell = \mathbf{h}_\ell^{\text{dyn}} + \mathbf{h}_\ell^{\text{stat}}.
\end{equation}

Each expert follows a LoRA formulation:
\begin{equation}
	f(\mathbf{z}) = \mathbf{B}\mathbf{A}\mathbf{z}, \quad 
	\mathbf{A}\in\mathbb{R}^{r\times d},\;
	\mathbf{B}\in\mathbb{R}^{d\times r},\; r\ll d.
\end{equation}

The experts are decomposed into two complementary categories: Dynamic experts ($\mathbf{h}_\ell^{\text{dyn}}$) that adaptively specialize in different missing patterns through task-aware routing, and Static experts ($\mathbf{h}_\ell^{\text{stat}}$) comprising shared experts for modality-invariant knowledge and modality-specific experts for individual modality priors. This design realizes the structured factorization in Eq.~(\ref{eq:factor}), enabling stable yet flexible missing-modality learning.

To investigate whether modality-missing learning can be modeled as multi-task adaptation, we compare MaMOL with representative MoE variants as illustrated in Fig.~\ref{fig:3}: Baseline Feed-Forward without experts, Adapt-Based MoE with lightweight adaptation branches, and Task-Driven MoE with task-embedding-based routing.

\subsection{Dynamic Experts: Pattern-aware Routing}
\label{sec:dynamic}

We introduce $K_d$ dynamic experts to model missing-pattern-induced adaptations:
\begin{equation}
	\mathbf{h}_\ell^{\text{dyn}} = \sum_{k=1}^{K_d} g_k^\ell(\mathbf{z}, \mathbf{m}) 
	\mathbf{B}_{k}^{\ell}\mathbf{A}_{k}^{\ell}\mathbf{z}.
\end{equation}

Unlike conventional MoE layers that depend solely on input features, our router explicitly incorporates both the modality-missing type and the semantic content:
\begin{equation}
	\mathbf{g}^\ell = \mathrm{TopK}\Big(\mathrm{softmax}(W_\ell \phi([\mathbf{z}; \psi(\mathbf{m})]))\Big),
\end{equation}
where $\phi(\cdot)$ is a lightweight projection network, $W_\ell$ is a trainable routing matrix, and $\psi(\mathbf{m})$ encodes the missing-modality configuration. Only the top-$K$ experts with the highest gating weights are selected, ensuring sparse activation and efficient computation.

Each dynamic expert approximates:
\begin{equation}
	f_k^\ell \approx \mathbb{E}[\Delta_\ell(\mathbf{z}) \mid \mathbf{m}=\mathbf{m}_k],
\end{equation}
allowing specialization over missing patterns while maintaining smooth knowledge sharing across related tasks.

\subsection{Static Experts: Shared and Modality-specific Experts}
\label{sec:static}

Static experts provide stable priors that complement the adaptive dynamic experts. We define two types with fixed activation coefficients:

A single shared expert ($\alpha_{\text{shr}} = 1$) is always activated to encode modality-invariant knowledge common across all missing-modality scenarios. This expert captures general visual representations and ensures stable performance across diverse configurations.

For each modality $i \in \{1, \dots, M\}$, a dedicated expert is activated only when that modality is present ($\alpha_i = m_i$). These experts adapt the pretrained backbone to remote sensing modalities and preserve modality-dependent priors, stabilizing feature alignment when certain modalities are absent.

Formally, the static expert output is:
\begin{equation}
	\mathbf{h}_\ell^{\text{stat}} = \underbrace{\mathbf{B}_{\text{shr}}^{\ell}\mathbf{A}_{\text{shr}}^{\ell}\mathbf{z}}_{\text{shared expert}} 
	+ \sum_{i=1}^{M} m_i \underbrace{\mathbf{B}_{i}^{\ell}\mathbf{A}_{i}^{\ell}\mathbf{z}}_{\text{modality-specific experts}}.
\end{equation}

Together with dynamic experts, this design realizes the structured factorization in Eq.~(\ref{eq:factor}), enabling stable yet adaptive missing-modality learning.

\begin{table*}[htbp]
	\centering
	\small  
	\caption{Performance comparison across datasets under different missing rates and train/test splits.}
	\label{tab:performance_all_1}
	\begin{adjustbox}{width=\textwidth}
		\renewcommand{\arraystretch}{.95}
		\setlength{\tabcolsep}{3pt} 
		\begin{tabular}{
				c c c
				*{6}{>{\centering\arraybackslash}p{2cm}} 
			}
			\toprule
			\multirow{2}{*}{\textbf{Dataset}} & \multirow{2}{*}{\textbf{Missing Rate}} & \multirow{2}{*}{\makecell[c]{\textbf{Train/Test}\\\textbf{HS other}}} &
			\multicolumn{6}{c}{\textbf{Method (OA\%/Kappa$\times$100)}} \\ 
			\cmidrule(lr){4-9}
			& & & \textbf{MMP} & \textbf{DCP} & \textbf{\textit{Moe\textsubscript{rep}}} & \textbf{\textit{Moe\textsubscript{ada}}} & \textbf{\textit{Moe\textsubscript{task}}} & \textbf{MaMOL} \\ 
			\midrule
			
			\multirow{9}{*}{\makecell[c]{\textbf{Houston13}\\(HS+LiDAR)}} 
			& \multirow{3}{*}{50\%} & 100\% 50\% & 91.56/90.95 & \underline{97.60/97.43} & 96.93/96.71 & 97.47/97.29 & 96.58/96.33 & \textbf{98.40/98.29} \\ 
			& & 50\% 100\% & 58.16/55.17 & 83.84/82.69 & \underline{85.84/84.83} & 77.80/76.21 & 84.96/83.88 & \textbf{88.00/87.14} \\ 
			& & 75\% 75\%  & 74.89/73.10 & 86.87/85.93 & \underline{87.27/86.36} & 81.36/80.02 & 86.36/85.38 & \textbf{93.87/93.43} \\ 
			\cmidrule(lr){2-9}
			& \multirow{3}{*}{70\%} & 100\% 30\% & 91.49/90.88 & \underline{98.11/97.98} & 96.80/96.57 & 96.09/95.81 & 97.02/96.81 & \textbf{98.24/98.12} \\ 
			& & 30\% 100\% & 44.42/40.45 & 79.24/77.76 & \underline{83.88/82.19} & 69.20/67.00 & 83.87/82.71 & \textbf{88.80/88.00} \\ 
			& & 65\% 65\%  & 65.51/63.05 & 85.64/84.62 & 85.16/84.10 & 85.51/84.48 & \underline{85.89/84.88} & \textbf{90.42/89.74} \\ 
			\cmidrule(lr){2-9}
			& \multirow{3}{*}{90\%} & 100\% 10\% & 90.89/90.24 & \underline{97.82/97.67} & 97.02/97.02 & 96.58/96.33 & 96.98/96.76 & \textbf{98.62/98.52} \\ 
			& & 10\% 100\% & 41.36/37.17 & 76.73/75.07 & 75.40/73.64 & 61.80/59.07 & \underline{83.44/82.26} & \textbf{85.69/84.67} \\ 
			& & 55\% 55\%  & 59.58/56.69 & 81.51/80.19 & \underline{82.20/80.93} & 79.49/77.37 & 81.00/79.64 & \textbf{87.33/86.43} \\ 
			\midrule
			
			\multirow{9}{*}{\makecell[c]{\textbf{Trento}\\(HS+LiDAR)}} 
			& \multirow{3}{*}{50\%} & 100\% 50\% & 90.39/88.47 & 97.50/97.00 & 97.33/96.80 & \underline{97.61/97.13} & 97.17/96.60 & \textbf{97.67/97.20} \\ 
			& & 50\% 100\% & 95.56/94.67 & 97.11/96.53 & \underline{97.44/96.93} & 96.94/96.33 & 97.39/96.87 & \textbf{98.06/97.67} \\ 
			& & 75\% 75\%  & 90.78/88.93 & \underline{96.39/95.67} & 95.67/94.80 & 95.50/94.60 & 96.06/95.27 & \textbf{98.39/98.07} \\ 
			\cmidrule(lr){2-9}
			& \multirow{3}{*}{70\%} & 100\% 30\% & 90.17/88.20 & \underline{97.67/97.20} & 97.39/96.87 & 96.44/95.73 & 96.89/96.27 & \textbf{97.78/97.33} \\ 
			& & 30\% 100\% & 93.78/92.53 & 96.44/95.73 & 96.89/96.27 & 96.72/96.07 & \underline{97.22/96.68} & \textbf{98.28/97.93} \\ 
			& & 65\% 65\%  & 88.44/86.13 & 94.94/93.93 & \underline{95.83/95.00} & 94.50/93.40 & 95.39/94.47 & \textbf{97.39/96.87} \\ 
			\cmidrule(lr){2-9}
			& \multirow{3}{*}{90\%} & 100\% 10\% & 86.50/83.80 & 96.98/96.20 & \underline{97.56/97.07} & 97.22/96.67 & 97.28/96.73 & \textbf{97.61/97.13} \\ 
			& & 10\% 100\% & 91.56/89.87 & 96.06/95.27 & 97.11/96.53 & \underline{97.33/96.80} & 97.39/96.87 & \textbf{97.44/96.93} \\ 
			& & 55\% 55\%  & 90.61/88.73 & \underline{95.89/95.07} & 94.33/93.20 & 94.39/93.27 & 95.78/94.93 & \textbf{95.94/95.13} \\ 
			\midrule
			
			\multirow{9}{*}{\makecell[c]{\textbf{Augsburg}\\(HS+SAR)}} 
			& \multirow{3}{*}{50\%} & 100\% 50\% & 58.05/51.06 & 78.57/75.00 & 77.10/73.28 & 73.48/69.06 & \textbf{79.52/76.11} & \underline{79.05/75.56} \\ 
			& & 50\% 100\% & 35.81/25.11 & \underline{43.43/34.00} & 43.05/33.56 & 42.48/32.89 & 42.86/33.33 & \textbf{44.62/35.39} \\ 
			& & 75\% 75\%  & 46.29/37.33 & \underline{61.29/54.83} & 59.48/52.72 & 58.52/51.61 & 58.62/51.72 & \textbf{62.86/56.67} \\ 
			\cmidrule(lr){2-9}
			& \multirow{3}{*}{70\%} & 100\% 30\% & 58.90/52.06 & 79.24/75.78 & 77.76/74.06 & 73.05/68.56 & \textbf{79.95/76.61} & \underline{79.90/76.56} \\ 
			& & 30\% 100\% & 26.33/14.06 & \underline{38.86/28.67} & 29.95/18.28 & 28.71/16.83 & 28.48/16.56 & \textbf{49.24/40.78} \\ 
			& & 65\% 65\%  & 44.05/34.72 & \underline{54.24/46.61} & 53.48/45.72 & 53.81/46.11 & 52.81/44.94 & \textbf{60.76/54.22} \\ 
			\cmidrule(lr){2-9}
			& \multirow{3}{*}{90\%} & 100\% 10\% & 60.05/53.39 & \textbf{79.48/76.06} & \underline{79.42/76.00} & 78.14/74.50 & 79.24/75.78 & 78.62/75.06 \\ 
			& & 10\% 100\% & 16.86/3.00 & \underline{35.00/24.17} & 17.90/4.22 & 17.67/3.94 & 17.95/4.28 & \textbf{42.00/32.33} \\ 
			& & 55\% 55\%  & 38.57/28.33 & \underline{47.48/38.72} & 45.90/36.89 & 46.57/37.67 & 45.81/36.78 & \textbf{56.00/48.67} \\ 
			\bottomrule
		\end{tabular}
	\end{adjustbox}
\end{table*}

\section{Experiments}

\subsection{Experimental Setup}
\paragraph{Datasets Description}
We evaluate MaMOL on three publicly available multimodal remote sensing datasets. Houston2013 \cite{contest2013ieee} provides hyperspectral (144 bands) and LiDAR data with 15 classes over the University of Houston campus. Augsburg \cite{baumgartner2012characterisation} includes hyperspectral (180 bands), SAR (4 polarization features), and LiDAR data from Augsburg, Germany, with 7 classes. Trento \cite{7902153} contains co-registered hyperspectral (63 bands) and LiDAR data acquired over an agricultural area near Trento, Italy, covering 6 land-cover classes.
\paragraph{Metrics}
We evaluate our method using Overall Accuracy (OA) and kappa across all datasets. OA measures the ratio of correctly predicted samples to the total number of samples, while kappa evaluates the agreement between predicted and true labels by accounting for the possibility of random chance, thus offering a more robust measure of classification reliability.
%

\paragraph{Implementation Details}
We adopt CLIP~\cite{Radford2021LearningTV} as the multimodal backbone, using ViT-B/16 as the visual encoder. All encoder parameters are frozen, and only the modality projection layers, lightweight expert modules, and the task-specific fully connected layer are trained. We set the number of dynamic experts to 2 with top-$k=1$ routing, and insert them into six designated Transformer layers. The model is optimized using Adam with a learning rate of $2 \times 10^{-3}$ and weight decay of $2 \times 10^{-2}$. A linear warm-up is applied over the first 10\% of training steps, followed by linear decay. All experiments are conducted on two NVIDIA RTX 3090 GPUs with a batch size of 128. For missing modalities, the corresponding encoder outputs are replaced with zero-filled tensors.

\paragraph{Setating of Missing Modality}
In this work, we consider realistic multimodal remote sensing scenarios, where the availability of different modalities may vary during both training and testing. To emulate such conditions, we define the \textit{missing rate}~$\eta$ as the ratio of modality-incomplete samples to the total number of samples, following the general formulation in MMP~\cite{lee2023cvpr}. Depending on which modalities are absent, several missing configurations can occur, including single-modality absence (only LiDAR missing). In the case of two missing modalities, the dataset is composed of $\frac{\eta}{2}$ samples missing the first modality, $\frac{\eta}{2}$ samples missing the second modality, and $(1-\eta)$ complete samples. For single-modality missing situations, $\eta$ of the data lacks one modality, while the remaining $(1-\eta)$ samples retain all modalities. This formulation can be naturally generalized to datasets with $M$ modalities, where each missing configuration occupies $(\frac{\eta}{M^2 - 2})$ of the data, and $(1-\eta)$ remains modality-complete.

\begin{figure*}[t]
	\centering
	\includegraphics[width=\linewidth]{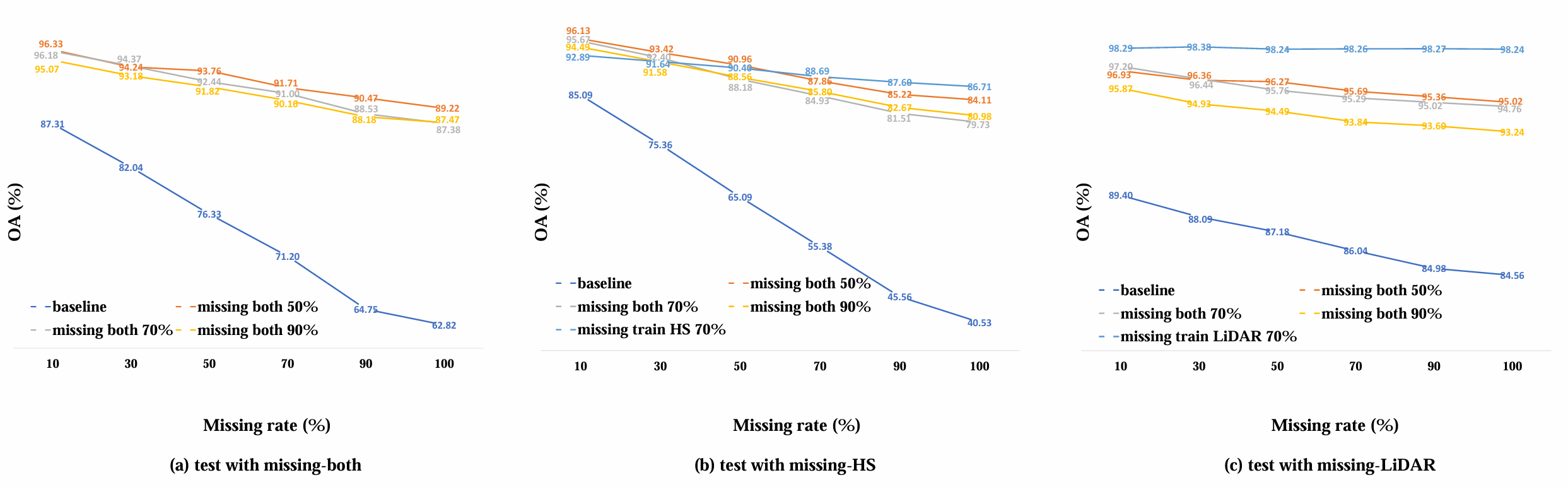}
	\caption{
		\textbf{Ablations on the generalizability of MaMOL across different testing scenarios under various missing rates on the Houston2013 dataset.} 
		(a) Models trained and tested on missing-both cases with different missing rates. 
		(b) Models trained on missing-both or missing-image cases and evaluated on missing-image conditions. 
		(c) Models trained on missing-both or missing-text cases and evaluated on missing-text conditions. 
	}
	\label{fig:generalizability}
\end{figure*}

\begin{table*}[t]
	\centering
	\small  
	\caption{Performance comparison across datasets under different train/test splits for natural image tasks.}
	\label{tab:performance_natural_image}
	\begin{tabular}{
			c c
			*{6}{c}  
		}
		\toprule
		\multirow{2}{*}{\textbf{Dataset}} &\multirow{2}{*}{\makecell[c]{\textbf{Train/Test}\\\textbf{Image Text}}} &
		\multicolumn{6}{c}{\textbf{Method (F1-Macro)}} \\ 
		\cmidrule(lr){3-8}
		& & \textbf{MMP} & \textbf{DCP} & \textbf{\textit{Moe\textsubscript{rep}}} & \textbf{\textit{Moe\textsubscript{ada}}} & \textbf{\textit{Moe\textsubscript{task}}} &  \textbf{MaMOL-3.6M}  \\ 
		\midrule
		
		\multirow{3}{*}{\makecell[c]{MM-IMDb}} 
		& 100\%  50\% & 40.67 & 49.25 & 38.14 & 40.91 & 42.01 & \textbf{51.73} \\ 
		& 50\%  100\% & 45.27 & 51.95 & 32.89 & 44.70 & 49.67 & \textbf{55.54} \\ 
		& 75\%  75\%  & 41.59 & 50.77 & 41.74 & 41.19 & 44.78 & \textbf{53.20} \\ 
		\midrule
		
		\bottomrule
	\end{tabular}
\end{table*}

\begin{table}[!t]
	\centering
	\small  
	\caption{Performance of MaMOL on Augsburg dataset (HS / SAR / LiDAR) under 75\% overall missing rate.}
	\label{tab:three_modal_missing75}
	\setlength{\tabcolsep}{6pt}
	\begin{tabular}{c c c c c}
		\toprule
		\textbf{Missing rate} & \textbf{HS} & \textbf{SAR} & \textbf{LiDAR} & \textbf{OA / Kappa$\times$100} \\
		\midrule
		\multirow{4}{*}{75\%} 
		& 75\% & 75\% & 75\%  & 80.10/76.78 \\ 
		& 100\% & 75\% & 50\%   & 83.29/80.50 \\ 
		& 50\% & 100\%  & 75\%  & 77.38/76.11 \\ 
		& 75\% & 50\%  & 100\%  & 67.19/61.72 \\ 
		\bottomrule
	\end{tabular}
\end{table}

\paragraph{Setting of MoE-based Methods}
We construct several MoE-based variants upon the ViLT baseline to investigate different expert integration strategies under missing-modality conditions, including replacement-based MoE~\cite{2022deepspeed}, adapter-based MoE~\cite{moe_lora_2024}, and task-guided MoE~\cite{liu2024moe}. All methods share the same backbone and training configuration to ensure fair comparison.

\subsection{Experimental Results}

\paragraph{Comparison with the State-of-the-arts}
We evaluate MaMOL on three widely used multimodal remote sensing datasets: Houston2013, Trento, and Augsburg, under various missing-modality scenarios. We compare MaMOL with MoE-based variants (\textit{Moe\textsubscript{rep}}, \textit{Moe\textsubscript{ada}}, \textit{Moe\textsubscript{task}}) and state-of-the-art methods (MMP, DCP), considering three missing rates (50\%, 70\%, 90\%) and three conditions: missing hyperspectral data, missing another modality, and missing both.

As shown in Table~\ref{tab:performance_all_1}, MaMOL consistently achieves the best performance across almost all settings, demonstrating its effectiveness in dynamic routing and expert selection under missing modalities. Although MoE-based baselines show improved robustness over prior methods, MaMOL exhibits clear advantages, especially on the Augsburg dataset where large modality discrepancies exist. These results validate the benefit of MaMOL’s missing-aware routing mechanism in handling severe and heterogeneous missing-modality conditions.

\paragraph{Robustness to Natural Image Task}
We further evaluate MaMOL on the MM-IMDb dataset to verify its generalization to natural image tasks. Comparisons are conducted against MMP, DCP, and three MoE-based variants under three train/test splits, using F1-Macro as the evaluation metric. As reported in Table~\ref{tab:performance_natural_image}, MaMOL consistently outperforms all baselines across all settings, confirming its robustness beyond remote sensing scenarios and its effectiveness for general multimodal classification tasks.

\paragraph{Modality Extending}
We additionally extend MaMOL to a three-modality missing setting on the Augsburg dataset (HS, SAR, LiDAR). Results in Table~\ref{tab:three_modal_missing75} show that MaMOL scales effectively to more complex multimodal configurations and achieves further performance gains, demonstrating its low-cost extensibility and strong robustness under multi-modality missing conditions.

\begin{table}[t]
	\centering
	\caption{Comparison of different MoE variants and expert layer configurations on Trento (70\% missing-both). Results are reported as Overall Accuracy (OA\%) / Kappa$\times$100.}
	\label{tab:method_comparison}
	\begin{adjustbox}{width=\linewidth}
		\renewcommand{\arraystretch}{1.15}
		\begin{tabular}{lcccc}
			\toprule
			\textbf{Method} 
			& \textbf{Baseline} 
			& \textbf{\textit{Head} (6×2)} 
			& \textbf{\textit{Last} (6×2)} 
			& \textbf{\textit{Last} (2×5)} \\ 
			\midrule
			\textit{Moe\textsubscript{rep}} & ViLT & 95.67 / 94.80 & \textbf{95.83 / 95.00} & 94.78 / 93.73 \\ 
			\textit{Moe\textsubscript{ada}} & ViLT & \textbf{95.50 / 94.60} & 94.50 / 93.40 & 94.67 / 93.60 \\ 
			\textit{Moe\textsubscript{task}} & ViLT & \textbf{96.44 / 95.73} & 95.39 / 94.47 & 94.56 / 93.47 \\ 
			\textbf{\textit{MaMOL\textsubscript{dyn}}} & CLIP & 96.11 / 95.33 & \textbf{97.39 / 96.87} & 96.94 / 96.33 \\ 
			\bottomrule
		\end{tabular}
	\end{adjustbox}
\end{table}

\begin{table}[t]
	\centering
	\caption{Ablation study on the trento (HS+LiDAR) dataset under different missing rates (OA\%).}
	\label{tab:ablation_houston13}
	\begin{adjustbox}{width=\linewidth}
		\renewcommand{\arraystretch}{1.1}
		\setlength{\tabcolsep}{6pt}
		\begin{tabular}{lcccc}
			\toprule
			\textbf{Method} 
			& \textbf{50\%} 
			& \textbf{70\%} 
			& \textbf{90\%} 
			& \textbf{Avg.} \\ 
			\midrule
			Baseline & 95.50 & 93.56 & 92.50 & 93.85 \\
			w/o Dynamic Experts & 96.72 & 96.39 & 94.22 & 95.78 \\
			w/o Static Experts & 97.56 & 96.78 & 95.06 & 96.47 \\
			w/o Modality-specific Experts & 97.67 & 97.17 & 95.28 & 96.71 \\
			\textbf{MaMOL (Ours)} & \textbf{98.39} & \textbf{97.39} & \textbf{95.94} & \textbf{97.24} \\
			\bottomrule
		\end{tabular}
	\end{adjustbox}
\end{table}

\begin{figure}[t]
	\centering
	\includegraphics[width=\linewidth]{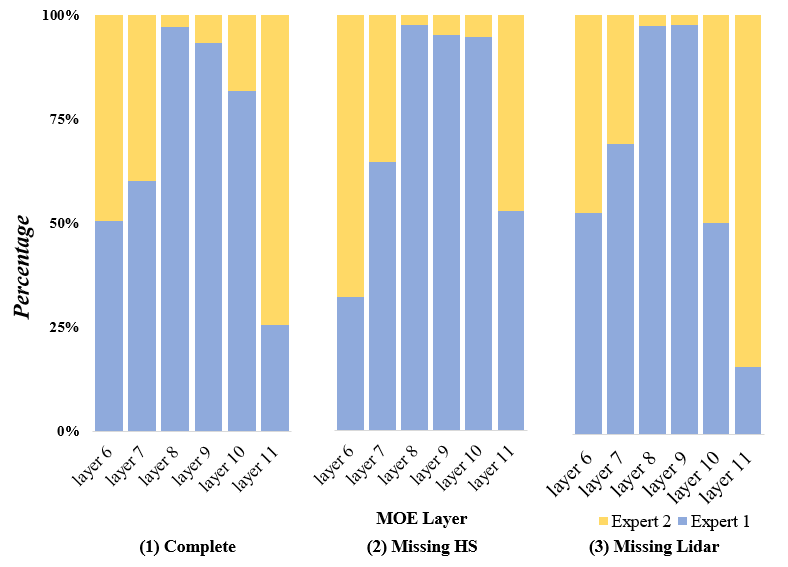}
	\caption{
		Layer-wise expert activation distributions under different modality-missing patterns on Houston2013. Statistics are computed on samples from \textbf{class 7} of the Houston2013 dataset.
	}
	\label{fig:routing_analysis}
\end{figure}

\subsection{Ablation Study}

\paragraph{Generalizability}
We evaluate the generalization ability of MaMOL on Houston2013 under mismatched training/testing missing-modality settings (Fig.~\ref{fig:generalizability}), including cross-condition transfer between missing-both, missing-image, and missing-text scenarios. MaMOL consistently outperforms the baseline across all configurations and missing rates, demonstrating strong robustness to unseen missing patterns. Models trained under specific missing conditions show slightly better performance on corresponding test cases, while maintaining stable behavior across diverse inputs. These results indicate that MaMOL generalizes well to complex and variable missing-modality scenarios.

\paragraph{Different Layers of Experts}
We study the impact of expert placement by inserting MoE modules at different transformer depths. As shown in Table~\ref{tab:method_comparison}, \textit{MaMOL\textsubscript{dyn}} consistently outperforms all ViLT-based MoE baselines under all configurations, achieving the best result when experts are placed in the last six layers. This confirms that dynamic routing in deeper layers is particularly effective for modeling high-level cross-modal interactions. We therefore adopt the “Last (6×2)” configuration in all subsequent experiments.

\paragraph{Effectiveness of Different Type Experts}
We conduct ablations on Trento under missing rates of 50\%, 70\%, and 90\% to analyze the contribution of each expert type (Table~\ref{tab:ablation_houston13}). Removing dynamic experts causes a substantial performance drop, highlighting their central role in adapting to diverse missing patterns. Removing static experts leads to reduced stability and accuracy, indicating their importance in preserving modality-invariant knowledge. Excluding modality-specific experts also degrades performance, confirming their contribution to modality-aware discrimination. The full MaMOL consistently achieves the best results, validating the complementary design of the three expert types.

\paragraph{Expert Routing Behavior Analysis}
We visualize layer-wise expert activation under complete inputs, missing-HS, and missing-LiDAR settings on Houston2013 (Fig.~\ref{fig:routing_analysis}). Expert utilization differs significantly across missing conditions, with discrepancies becoming more pronounced in deeper layers. This shows that MaMOL dynamically modulates high-level representations according to modality availability. Moreover, multiple experts are activated even under complete inputs, indicating that MaMOL captures diverse cross-modal interaction patterns. These results confirm that MaMOL enables missing-aware conditional computation.

\section{Conclusion}
\label{sec:conclusion}

In this paper, we propose Missing-aware Mixture-of-Loras(MaMOL), a unified and efficient framework for multimodal classification under modality-missing conditions. MaMOL reformulates modality missing as an expert selection problem from a mixture-of-experts perspective. By combining task-oriented dynamic experts and modality-specific–shared static experts, it enables flexible adaptation to diverse missing patterns while maintaining cross-modal consistency. Extensive experiments on multiple remote sensing datasets demonstrate the superior performance and robustness of MaMOL and other MoE-based variants under various missing rates and modality combinations. MaMOL also generalizes well to more complex three-modality scenarios and even to natural image tasks, showing strong scalability and cross-domain generalization. Overall, MaMOL offers a lightweight, robust, and extensible solution for real-world multimodal incompleteness, providing new insights toward building missing-aware and generalizable multimodal learning systems.

\nocite{langley00}

\bibliography{example_paper}
\bibliographystyle{icml2026}

\newpage
\appendix
\onecolumn

\end{document}